\ifdefined\ARXIVVERSION
\documentclass{article}
\usepackage{arxiv}
\usepackage[utf8]{inputenc}
\usepackage[T1]{fontenc}
\usepackage{hyperref}
\usepackage{natbib}
\usepackage{doi}
\usepackage{microtype}
\else
\documentclass[conference]{IEEEtran}
\IEEEoverridecommandlockouts
\usepackage{cite}
\fi

\usepackage{amsmath,amssymb,amsfonts}
\usepackage{booktabs}
\usepackage{graphicx}
\usepackage{textcomp}
\usepackage{xcolor}
\usepackage{xspace}
\usepackage{url}

\newcommand{\method}{\textsc{EARM}\xspace}

\newcommand{\R}{\mathbb{R}}

\begin{document}

\title{The Retriever Should Remember:\\
Experience-Amortized Reranking for Long-Term Agent Memory}

\ifdefined\ARXIVVERSION
\author{
    Qi Feng \qquad Chris Ding \qquad Jicong Fan\thanks{Corresponding author}\\
    School of Data Science\\
    The Chinese University of Hong Kong, Shenzhen\\
    \texttt{qifeng@link.cuhk.edu.cn} \quad \texttt{chrisding@cuhk.edu.cn}\\
    \texttt{fanjicong@cuhk.edu.cn}
}
\date{}
\renewcommand{\shorttitle}{The Retriever Should Remember}
\hypersetup{
    pdftitle={The Retriever Should Remember: Experience-Amortized Reranking for Long-Term Agent Memory},
    pdfauthor={Qi Feng, Chris Ding, Jicong Fan},
    pdfkeywords={agent memory, long-term memory, retrieval, LLM reranking, matrix completion, amortized inference}
}
\else
\author{    Qi Feng \qquad Chris Ding \qquad Jicong Fan\thanks{Corresponding author}\\
    School of Data Science\\
    The Chinese University of Hong Kong, Shenzhen\\
    \texttt{qifeng@link.cuhk.edu.cn} \quad \texttt{chrisding@cuhk.edu.cn}\\
    \texttt{fanjicong@cuhk.edu.cn}}
\fi

\maketitle

\begin{abstract}
Long-term language-model agents accumulate memories across interactions, but their retrievers typically do not accumulate retrieval experience.
Semantic retrieval is efficient, but embedding similarity does not always reflect whether a memory contains evidence relevant to the current query.
Large language model (LLM) rerankers provide stronger query-conditioned relevance scores, yet stateless reranking repeatedly scores a large candidate pool and discards these scores after each query.
We introduce \method, an experience-amortized reranking framework that treats previously acquired LLM relevance scores as reusable retrieval experience.
\method stores sparse query--memory relevance scores in an online matrix, learns their shared structure through causal matrix completion, and combines a small set of newly observed scores with estimated scores to rerank the remaining candidates.
The scoring budget decreases as experience accumulates, changing LLM reranking from a repeated per-query expense into a retrieval capability learned over an agent's lifetime.
Experiments on long-term conversational memory show that mixed observed-and-estimated reranking improves answer accuracy over semantic retrieval by up to 6.62\% and remains effective when only 17.5\% of candidates receive direct LLM relevance scores, thereby substantially reducing the inference overhead of LLM reranking.
These results motivate a broader view of agent memory: a long-lived agent should remember not only past content, but also how that content has proved useful for retrieval. Our code is available at \url{https://github.com/FengQi-HITSZ/earm}.
\end{abstract}

\ifdefined\ARXIVVERSION
\else
\fi

\section{Introduction}
\label{sec:introduction}

Long-lived language-model agents must carry information across interactions that cannot remain in the immediate context~\cite{park2023generative,liu-etal-2024-lost}.
External memory makes this persistence possible by storing events, user preferences, task trajectories, and other reusable information~\cite{zhong2024memorybank,packer2023memgpt,chhikara2025mem0,xu2025amem}.
Storage alone, however, does not make an agent remember effectively.
As the memory store grows, the system must repeatedly identify a small set of entries that supplies the evidence needed by the current query.
Retrieval therefore becomes part of the memory mechanism itself: a stored experience can affect the agent only if it is surfaced at the right time.

Many current systems, however, exhibit a basic asymmetry: the memory store evolves as interactions accumulate, but the retrieval procedure starts from the same initial state for every query.
A typical system retrieves candidates using embedding similarity, optionally uses an LLM to assign a relevance score to each candidate, consumes the selected memories once, and then discards those scores.
The same memory may later appear in another candidate set, and a related information need may recur, but the retriever starts over.
The agent therefore accumulates \emph{content memory} without accumulating experience about how that content should be accessed.
Our starting observation is simple: \emph{agent memory systems are designed to remember past interactions, yet their retrievers forget every retrieval they perform}.

Semantic retrieval alone does not resolve this problem.
Dense embedding similarity provides an efficient first-stage filter~\cite{karpukhin-etal-2020-dense}, but representational similarity is not equivalent to query-conditioned memory relevance.
A memory can be lexically or topically close to a query while adding no evidence needed for the answer; an apparently distant memory can instead supply a missing temporal, causal, or entity relation.
Relevance can also depend on the current dialogue context and on how a memory complements other selected evidence.
These conditions explain why a high-similarity candidate pool can still contain substantial noise and why its original order can be a poor answer-oriented ranking~\cite{pan2025secom,yoran2024robust}.
Recent studies reach the same conclusion from complementary perspectives: similarity-based access degrades as relevant evidence becomes temporally distant, and a semantically related memory can still be inappropriate for the current task~\cite{zhou2026agent-native,zhang2026beyond-similarity}.

LLM rerankers reduce this mismatch by jointly interpreting the query and a candidate memory.
Prior work shows that language models can serve as effective text rankers~\cite{sachan-etal-2022-improving,sun-etal-2023-chatgpt,qin-etal-2024-large}, including under explicit inference budgets~\cite{rashid-etal-2024-ecorank}.
However, conventional reranking treats queries independently.
For $T$ queries with $N$ candidates each, pointwise reranking requires up to $TN$ LLM relevance-scoring operations.
This cost is especially problematic for long-lived agents because it grows with both the number of interactions and the size of the retrieved pool.
More importantly, the cost is repetitive: each score reveals something about the accessibility of the agent's memory, but a stateless reranker does not preserve that evidence.

We reinterpret these LLM scores as \emph{retrieval experience}.
Across an agent's lifetime, the scores form a sparse relevance matrix whose rows are stable memory identities and whose columns are queries.
An observed entry records the relevance assigned to one query--memory pair by the LLM reranker.
The matrix can exhibit shared structure: some memories are relevant across several information needs; related queries can induce similar relevance patterns; and groups of memories and queries can interact through recurring latent factors.
We call this structure a \emph{latent memory accessibility map}.
If the map can be learned online, a new query needs only a small number of LLM-scored anchors to locate itself within the map; the system can estimate the relevance of the remaining candidates from accumulated experience.

Based on this view, we introduce \method (Experience-Amortized Reranking for Memory).
For each incoming query, \method first obtains a semantic candidate pool and asks the LLM to score only a budgeted subset.
The subset balances exploitation---sampling candidates from the high-similarity region---with exploration through random sampling from the remaining candidates.
\method inserts these scores into a causal query--memory matrix, updates a biased low-rank model with warm starts, and predicts the unobserved candidate scores.
The final ranking uses the observed LLM relevance score when available and the completed score otherwise.
Because the matrix also contains memories omitted by the current semantic retriever, completion can further turn that first-stage pool into a soft boundary: historically supported memories outside the pool can be scored and reconsidered when cosine similarity misses them.
During cold start, the system observes more scores; as retrieval experience accumulates, a stage-wise schedule reduces the observation budget.
The system thus pays an initial cost to learn how its own memory should be retrieved and amortizes this cost over its lifetime.

We evaluate the framework end to end by using sparse LLM relevance scores to rerank retrieved memories before answer generation.
Across the evaluated settings, mixed observed-and-completed rankings improve answer accuracy over semantic retrieval by 2.14--6.62\% and approach full LLM reranking to within 2.34--2.79\% in the strongest completion configurations.
The component ablation further shows that completed-score-selected memories add 0.78--2.79\% beyond the directly scored memories alone.

This paper makes two contributions:
\begin{itemize}
    \item We introduce \method, an experience-amortized reranking framework that maintains LLM relevance scores as persistent retrieval state and combines stratified anchor scoring, causal low-rank completion, and mixed observed--estimated ranking for a sequential query stream.
    \item We demonstrate that \method improves answer accuracy over semantic retrieval by up to 6.62\%, remains effective at a 17.5\% direct-scoring ratio, and obtains consistent additional gains from completed-score-selected memories beyond the observed anchors alone.
\end{itemize}

\section{Related Work}
\label{sec:related}

\subsection{Long-Term Agent Memory and Retrieval}

Long-term agent memory has evolved from persistent memory streams and tiered storage~\cite{park2023generative,zhong2024memorybank,packer2023memgpt} toward a lifecycle spanning extraction, representation, retrieval, and maintenance~\cite{zhang2025memory-survey,zhou2026agent-native}.
Recent systems extract and consolidate conversational facts or dynamically link memories~\cite{chhikara2025mem0,xu2025amem}, compress long histories or organize working memory hierarchically~\cite{lee2024readagent,hu-etal-2025-hiagent}, structure stores for efficient access~\cite{fang2026lightmem,liu2026simplemem}, learn associative organization~\cite{zhu-etal-2026-hela}, or delegate memory operations to smaller models~\cite{zhang-etal-2026-lightweight}.
Long-horizon benchmarks test whether these stores recover multi-hop, temporal, and cross-session evidence~\cite{maharana-etal-2024-evaluating,wu2025longmemeval,tan-etal-2025-membench}.
Retrieval methods therefore enrich memory graphs and propagate activation across them~\cite{gutierrez2024hipporag,chhikara2025mem0,xu2025amem,jiang2026synapse}, or learn query-conditioned admission beyond embedding similarity~\cite{zhang2026beyond-similarity}.
\method keeps extraction and storage fixed and instead makes query--memory relevance scores persistent across retrievals.

\subsection{Experience-Amortized Reranking}

Other agents reuse past trajectories by storing verbal reflections, distilling transferable lessons, or inducing reusable workflows for subsequent tasks~\cite{shinn2023reflexion,zhao2024expel,wang2025awm}.
MemRL also makes retrieval improve from experience by updating memory utilities from environmental feedback and selecting episodic strategies with learned Q-values~\cite{zhang2026memrl}.
\method differs in both signal and prediction target: it stores pointwise LLM relevance scores for query--memory pairs and predicts unseen pairs through their shared structure, without requiring task rewards or assigning one scalar utility to each memory.
LLM rerankers provide this stronger query-conditioned signal by jointly interpreting queries and candidates~\cite{qin-etal-2024-large}, while budget-constrained rerankers reduce the inference spent on the current query~\cite{rashid-etal-2024-ecorank}.
In contrast, \method amortizes scoring across a persistent memory store and a sequential query stream.
Its bias-aware low-rank completion builds on matrix factorization~\cite{koren2009matrix}, while its growing-column, fixed-observation-budget setting is related to budgeted column-space recovery~\cite{kim2020recommendation}.
Unlike this prior formulation, the entries are acquired through LLM inference, only causal history is available, and the objective is downstream answer quality rather than matrix reconstruction itself.

\section{Problem Formulation}
\label{sec:problem}

Consider a chronological query stream $q_1,\ldots,q_T$ and a growing memory store $\mathcal{M}_t$ whose entries retain stable identities once written.
For query $q_t$, a first-stage semantic retriever returns a candidate set $C_t\subseteq\mathcal{M}_t$ with $|C_t|=N$.
An LLM relevance scorer can then evaluate each query--memory pair:
\begin{equation}
    s^{\mathrm{LLM}}_{it}=g_{\phi}(q_t,m_i)\in[0,1],
    \qquad m_i\in C_t,
    \label{eq:llm-score}
\end{equation}
where a higher score indicates that $m_i$ is more relevant to $q_t$ according to the reranker.
A full-reranking policy evaluates Eq.~\eqref{eq:llm-score} for all $N$ candidates independently at every query.

Our setting permits only $B_t<N$ relevance-scoring operations for query $q_t$.
The scorer is applied to an anchor subset $O_t\subseteq C_t$, with $|O_t|=B_t$.
The system must use the observed scores in $O_t$ together with relevance scores accumulated from earlier queries to estimate the unscored candidates $C_t\setminus O_t$ and select $K$ memories for the answer context.
The task is therefore sequential sparse reranking: approximate the ranking produced by dense LLM relevance scoring while amortizing scoring effort across the query stream.

The formulation assumes that memory identities remain stable and that query--memory relevance contains sufficient recurring structure for past scores to inform future rankings.
Section~\ref{sec:limitations} discusses settings in which these assumptions fail.

\section{Experience-Amortized Memory Reranking}
\label{sec:method}

\begin{figure*}[t]
    \centering
    \includegraphics[width=\textwidth]{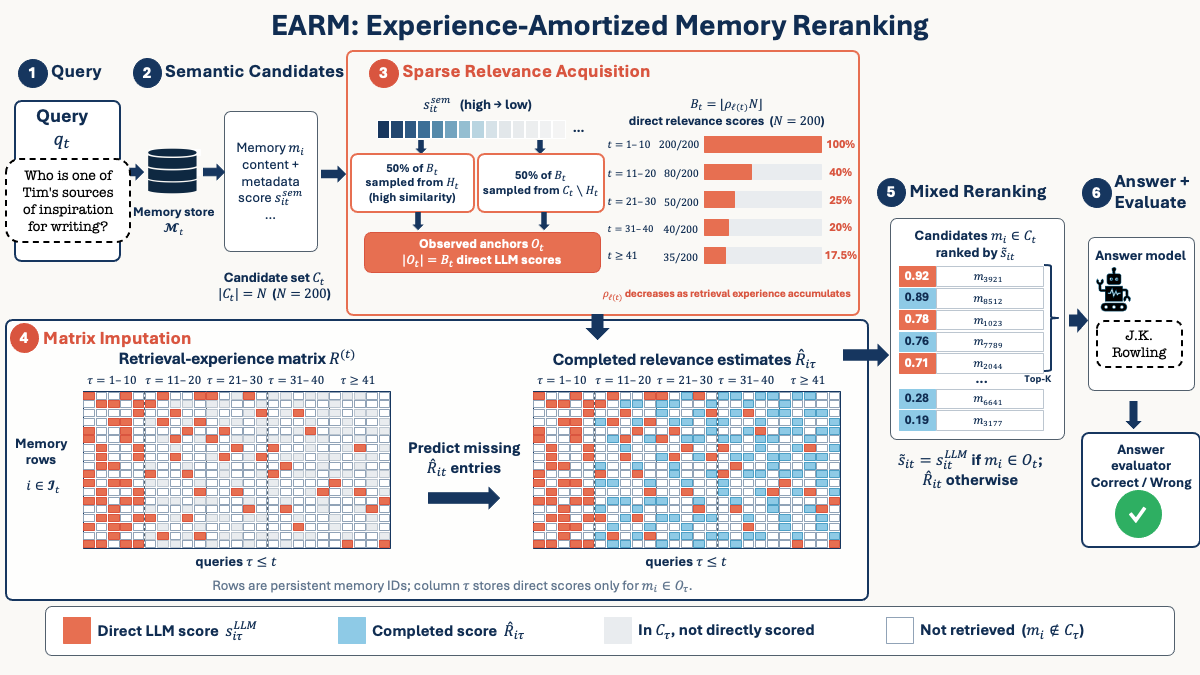}
    \caption{Overview of \method. Semantic retrieval forms a candidate set for each query; a budgeted subset receives direct LLM relevance scores; causal matrix completion estimates the missing query--memory scores; and mixed reranking selects the answer context using direct scores when observed and completed scores otherwise. White cells denote memories that were available but not retrieved for a query. Although the experiments rerank within the semantic candidate set, the same completion model can score historically supported white cells and thereby expand the search beyond cosine retrieval. Selected memories are restored to chronological order before answer generation and evaluation.}
    \label{fig:earm-pipeline}
\end{figure*}

\method converts sparse LLM relevance scores into a retrieval model that develops over the agent's lifetime.
Its five components are semantic candidate generation, experience acquisition, a causal retrieval-experience matrix, online low-rank completion, and mixed reranking.
Figure~\ref{fig:earm-pipeline} summarizes the complete pipeline from query arrival to answer evaluation.

\subsection{Semantic Candidate Generation}

For each query $q_t$, the semantic retriever computes an embedding similarity $s^{\mathrm{sem}}_{it}$ and returns the top $N$ memories.
This stage is deliberately high-recall: it narrows the full memory store to a manageable pool but does not determine the final context.
Candidate identities are preserved across queries, so a memory retrieved at several times always occupies the same matrix row.

\subsection{Retrieval Experience Matrix}

Let $\mathcal{I}_t=\bigcup_{\tau\leq t}C_{\tau}$ denote the set of memory entries that have appeared in at least one semantic candidate set up to query $t$.
We construct a sparse matrix $R^{(t)}\in\R^{|\mathcal{I}_t|\times t}$ with
\begin{equation}
    R^{(t)}_{i\tau}=
    \begin{cases}
        s^{\mathrm{LLM}}_{i\tau}, & m_i\in O_{\tau},\\
        \text{unobserved}, & \text{otherwise}.
    \end{cases}
    \label{eq:experience-matrix}
\end{equation}
Each column represents a query, each row represents a persistent memory entry, and each observed entry records an LLM relevance score already acquired by the system.
An unobserved cell can therefore arise in two ways: the memory was retrieved but not directly scored, or the memory was available in the store but omitted from that query's semantic candidate set.
These two cases correspond respectively to the gray and white cells in Figure~\ref{fig:earm-pipeline}; cells preceding a memory's insertion into a growing store are structurally unavailable and are excluded from completion.
We refer to $R^{(t)}$ as the \emph{retrieval-experience matrix} to distinguish it from a temporary per-query score list.

\subsection{Stage-Wise Experience Acquisition}
\label{sec:sampling}

We divide the query stream into experience stages.
Let $\ell(t)$ denote the stage containing query $t$, and let $\rho_{\ell(t)}\in(0,1]$ be its observation ratio.
The relevance-scoring budget is
\begin{equation}
    B_t=\left\lfloor\rho_{\ell(t)}N\right\rfloor,
    \qquad
    \rho_1\geq\rho_2\geq\cdots>0.
\label{eq:budget}
\end{equation}
The initial stage uses a larger ratio to establish the retrieval-experience matrix.
Later stages use progressively smaller ratios as more relevance structure becomes available.
Equation~\eqref{eq:budget} describes a predefined maturity schedule rather than an uncertainty-adaptive policy; the instantiated stage boundaries and ratios are experimental hyperparameters.

To construct $O_t$, we partition $C_t$ into a high-similarity stratum $H_t$ and its complement $C_t\setminus H_t$.
Half of the budget is randomly sampled from $H_t$, and half is randomly sampled from $C_t\setminus H_t$.
The former exploits the first-stage retriever by acquiring direct relevance scores in a promising region.
The latter explores lower-ranked regions, increases matrix coverage, and prevents the completion model from observing only the semantic retriever's preferred examples.
The equal split is fixed across stages; its effect can be isolated through a sampling-policy ablation.

\subsection{Causal Relevance Completion}

The retrieval-experience matrix is incomplete by construction: each query column contains direct scores for only its anchor subset.
We estimate the missing entries with a bias-aware low-rank model,
\begin{equation}
    \widehat{R}_{i\tau}=\mu+\alpha_i+\beta_{\tau}
    +p_i^{\top}z_{\tau},
    \label{eq:factorization}
\end{equation}
where $\mu$ is the global relevance level, $\alpha_i$ is a memory-specific bias, $\beta_{\tau}$ is a query-specific bias, and $p_i,z_{\tau}\in\R^r$ are latent memory and query factors.
The bias terms separate generally high-scoring memories and generally broad queries from pair-specific structure.
The interaction $p_i^{\top}z_{\tau}$ captures recurring relevance patterns shared across different query--memory pairs.

Let $\Omega_t=\{(i,\tau):\tau\leq t, m_i\in O_{\tau}\}$ be the entries observed by the time query $t$ is processed.
The model minimizes masked reconstruction error over $\Omega_t$ together with regularization:
\begin{equation}
    \mathcal{L}_t=
    \sum_{(i,\tau)\in\Omega_t}
    \left(R_{i\tau}-\widehat{R}_{i\tau}\right)^2
    +\mathcal{R}(\alpha,\beta,P,Z).
    \label{eq:completion-loss}
\end{equation}
Here, $\mathcal{R}(\alpha,\beta,P,Z)=\lambda\bigl(\|\alpha\|_2^2+\|\beta\|_2^2+\|P\|_F^2+\|Z\|_F^2\bigr)$ is an $L_2$ regularizer on the memory and query biases and their latent factors, with $\lambda$ controlling the regularization strength.
We optimize Eq.~\eqref{eq:completion-loss} with alternating least squares and warm-start each update from the previous solution.

The anchors in the current query column serve a specific role.
Historical columns estimate the shared memory parameters $\alpha_i$ and $p_i$.
The observed entries in $O_t$ then constrain the new column parameters $\beta_t$ and $z_t$.
Once the current query is positioned relative to the historical relevance structure, Eq.~\eqref{eq:factorization} estimates scores for $C_t\setminus O_t$.
The prediction domain need not stop at $C_t$.
For any memory $m_i\in\mathcal{I}_{t-1}\setminus C_t$ whose row parameters have been learned from earlier observations, Eq.~\eqref{eq:factorization} also estimates the corresponding white cell $\widehat{R}_{it}$.
Let $E_t\subseteq\mathcal{I}_{t-1}\setminus C_t$ denote such expansion memories and define the ranking domain $D_t=C_t\cup E_t$.
Including $E_t$ allows a memory omitted by the current cosine retriever to re-enter the search through its completed relevance score; memories with no historical row support remain cold-start items and require direct acquisition or an inductive initialization.
After processing the query, its observed anchors remain in the matrix and refine the shared parameters used by later queries.
This repeated update is the mechanism through which reranking effort is amortized across time.

The causal mask is essential.
When ranking query $q_t$, the model may use historical observations and the sampled anchors in column $t$, but it may not use scores from later queries.
The resulting factors represent a causal, evolving map of memory accessibility rather than an offline reconstruction of the complete test matrix.

\subsection{Mixed Reranking and Context Construction}

For each memory $m_i\in D_t$, \method uses
\begin{equation}
    \widetilde{s}_{it}=
    \begin{cases}
        s^{\mathrm{LLM}}_{it}, & m_i\in O_t,\\
        \widehat{R}_{it}, & m_i\in D_t\setminus O_t.
    \end{cases}
    \label{eq:mixed-score}
\end{equation}
The experiments use $D_t=C_t$ to evaluate sparse reranking under the same first-stage retrieval pool as the baselines.
The expanded form uses $E_t\neq\varnothing$ and turns completion into candidate expansion when cosine retrieval is an unreliable boundary.
Memories in $D_t$ are sorted by $\widetilde{s}_{it}$, and the highest-scoring $K$ memories form the answer context.
Selection order and presentation order serve different purposes: relevance scores determine \emph{which} memories enter the context, after which the selected memories are restored to chronological order to preserve the temporal coherence of the conversation.

This mixed policy retains direct LLM relevance scores where they are available while allowing completed scores to compete for the remaining context slots.
As the agent matures, observed scores act as anchors that locate a new query relative to the accessibility map learned from earlier interactions.

\section{Experimental Design}
\label{sec:experiments}

We organize the evaluation around three research questions:
\begin{enumerate}
    \item \textbf{RQ1:} How accurately does experience-amortized reranking answer different question types relative to semantic retrieval and full LLM reranking?
    \item \textbf{RQ2:} Does the advantage over semantic retrieval persist as the direct-scoring ratio decreases across experience stages?
    \item \textbf{RQ3:} How much of the overall improvement comes from directly scored memories, and how much is added by memories selected through completed scores?
\end{enumerate}

\subsection{Dataset and End-to-End Pipeline}

We evaluate on LoCoMo~\cite{maharana-etal-2024-evaluating}, which contains long conversations and questions that test long-term conversational memory.
We adopt the memory extraction and storage pipeline of Mem0~\cite{chhikara2025mem0} and modify only the retrieval and reranking stages.
For each conversation, we construct the complete memory store before processing its questions and keep it fixed throughout evaluation; hence $\mathcal{M}_t=\mathcal{M}$ for every experimental query index $t$.
Only the query columns and observed relevance scores in $R^{(t)}$ accumulate with $t$---the memory population itself does not grow across experience stages.
Questions within each conversation are processed in their dataset order.
For every query, the semantic retriever returns a fixed-size candidate set.
The current LLM relevance scorer is Qwen3.5-4B-Q8\_0~\cite{qwen2026qwen35}, prompted to return a relevance score in $[0,1]$ for a query--memory pair.
After reranking, the top $K\in\{10,20\}$ memories are ordered chronologically and provided to GPT-4o-mini, which generates the answer.
Separately, GPT-4o-mini~\cite{openai2024gpt4omini} serves as the LLM-based answer evaluator~\cite{zheng2023llmjudge} that compares the generated answer with the reference answer.

\subsection{Implementation Details}

The semantic retriever returns $N=200$ candidates per query.
Within each conversation, queries are grouped into blocks of ten.
The relevance-scoring budgets are 200, 80, 50, 40, and 35 candidates for query indices 1--10, 11--20, 21--30, 31--40, and 41 onward, corresponding to observation ratios of 100\%, 40\%, 25\%, 20\%, and 17.5\%.
When results are aggregated over the evaluated conversations, the first four stages contain 100 questions each; the final stage contains all remaining questions.
After the first block, the sampling policy assigns half of each budget to the high-similarity stratum and half to the remaining candidates.
The completion model uses $\ell_2$ regularization with coefficient $0.1$ and evaluates latent ranks $r\in\{2,8\}$.
ALS runs for at most 5,000 iterations and terminates when the improvement falls below $10^{-6}$.

\subsection{Baselines}

We compare four retrieval policies:
\begin{itemize}
    \item \textbf{Semantic}: retain the first-stage semantic order and apply no LLM reranker.
    \item \textbf{Full LLM reranking}: score all 200 candidates with the LLM relevance scorer and rank them without completion.
    \item \textbf{Observed-only}: disable score completion and place only directly scored memories among the top $K$ into the answer context.
    \item \textbf{Mixed (\method)}: rank all candidates using Eq.~\eqref{eq:mixed-score} and return the top $K$ memories.
\end{itemize}

\subsection{Metrics and Statistical Analysis}

The primary end-to-end metric is answer accuracy assigned by the LLM-based answer evaluator.
We report overall accuracy and accuracy on multi-hop, open-domain, single-hop, and temporal questions.
We count one LLM reranker call for each directly scored query--memory pair.
For the component ablation, we report absolute accuracy differences in percent (\%) and the average number of directly scored memories in the selected context.

\subsection{Comparison Logic}

RQ1 compares sparse completion with the two endpoints of the reranking spectrum: semantic retrieval uses no LLM relevance scores, whereas full LLM reranking scores every candidate.
RQ2 groups queries by the experience stages defined above and tests whether completion continues to improve over semantic retrieval when the direct-scoring ratio falls from 100\% to 17.5\%.
RQ3 disables completion while preserving the same observed anchors.
The Observed-only--Semantic difference measures the contribution of directly scored memories, while the Mixed--Observed-only difference measures the additional contribution of memories admitted by completed scores.
Because Semantic and Mixed both return the same compact context budget $K\in\{10,20\}$, the end-to-end Semantic--Mixed comparison holds the final context size fixed.
These compact budgets also limit the amount of distracting context supplied to the answer model~\cite{liu-etal-2024-lost,yoran2024robust}, making the ablation primarily sensitive to whether the selected memories contain the evidence needed for the query.

\section{Results}
\label{sec:results}

\subsection{Overall and Question-Type Accuracy}

Table~\ref{tab:main-results} compares semantic retrieval, \method, and full LLM reranking.
For Top-10 retrieval, rank-8 \method improves overall accuracy from 82.21\% to 88.83\%, a gain of 6.62\%, and closes the gap to full LLM reranking to 2.79\%.
The improvement is largest on multi-hop questions, where accuracy increases by 10.29\%, followed by single-hop questions with a 7.25\% increase.
For Top-20 retrieval, rank-2 \method performs best among the completion variants, improving overall accuracy by 2.99\% and remaining 2.34\% below full reranking.
Across the evaluation, \method makes 78,736 LLM reranker calls, compared with 307,982 for full reranking, reducing the total number of calls by 74.43\%.
With this reduction, the best \method configuration recovers approximately 70\% of the improvement offered by full reranking over Semantic at Top-10 and 56\% at Top-20.

\begin{table*}[h!]
\centering
\caption{Answer accuracy (\%) on LoCoMo by question type. Full LLM reranking scores all 200 candidates; \method uses the stage-wise sparse scoring schedule. Bold values are the best result for each Top-$K$ setting.}
\label{tab:main-results}
\resizebox{0.85\textwidth}{!}{%
\begin{tabular}{clccccc}
\toprule
Top-$K$ & Method & Overall & Multi-hop & Open-domain & Single-hop & Temporal \\
\midrule
10 & Semantic & 82.21 & 77.30 & 72.92 & 85.26 & 81.31 \\
10 & \method ($r=2$) & 87.27 & 87.59 & 73.96 & 91.44 & 80.06 \\
10 & \method ($r=8$) & 88.83 & 87.59 & 75.00 & 92.51 & 84.42 \\
10 & Full LLM reranking & \textbf{91.62} & \textbf{90.43} & \textbf{83.33} & \textbf{95.24} & \textbf{85.67} \\
\midrule
20 & Semantic & 87.08 & 87.94 & 78.13 & 89.18 & 83.49 \\
20 & \method ($r=2$) & 90.06 & 90.78 & 80.21 & 93.46 & 83.49 \\
20 & \method ($r=8$) & 89.22 & 90.43 & 78.13 & 92.15 & 83.80 \\
20 & Full LLM reranking & \textbf{92.40} & \textbf{93.62} & \textbf{82.29} & \textbf{94.41} & \textbf{89.10} \\
\bottomrule
\end{tabular}%
}
\end{table*}

Across question types, the gains are most pronounced on multi-hop and single-hop questions, while open-domain and temporal performance remains closer to semantic retrieval.
Overall, \method retains a substantial share of the answer-quality benefit of full reranking while using roughly one quarter of its LLM calls.

\subsection{Accuracy Across Experience Stages}

Table~\ref{tab:block-results} reports accuracy as the observation ratio decreases across query stages.
As the direct-scoring ratio falls from 100\% to 17.5\%, both completion ranks continue to outperform Semantic for both context sizes.
In B5+, where only 17.5\% of the candidate pairs receive an LLM relevance score, rank-8 \method improves Top-10 accuracy by 5.53\%, while rank-2 \method improves Top-20 accuracy by 2.46\%.

\begin{table*}[t]
\centering
\caption{Answer accuracy (\%) across experience stages. B1--B4 contain 100 aggregated questions each; B5+ contains the remaining questions. Percentages in parentheses are the fractions of the 200 candidates directly scored by the LLM relevance scorer. Answer generation and answer evaluation were run independently for each method, so small block-level fluctuations include generation and evaluation variance.}
\label{tab:block-results}
\resizebox{0.9\textwidth}{!}{%
\begin{tabular}{clcccccc}
\toprule
Top-$K$ & Method & B1 (100\%) & B2 (40\%) & B3 (25\%) & B4 (20\%) & B5+ (17.5\%) & Overall \\
\midrule
10 & Semantic & 76.00 & 78.00 & 76.00 & 77.00 & 84.12 & 82.21 \\
10 & \method ($r=2$) & 86.00 & 86.00 & 86.00 & 81.00 & 88.16 & 87.27 \\
10 & \method ($r=8$) & 88.00 & 86.00 & 86.00 & 86.00 & 89.65 & 88.83 \\
10 & Full LLM reranking & 85.00 & 90.00 & 88.00 & 88.00 & 92.98 & 91.62 \\
\midrule
20 & Semantic & 81.00 & 84.00 & 89.00 & 80.00 & 88.33 & 87.08 \\
20 & \method ($r=2$) & 88.00 & 87.00 & 91.00 & 86.00 & 90.79 & 90.06 \\
20 & \method ($r=8$) & 87.00 & 91.00 & 90.00 & 88.00 & 89.30 & 89.22 \\
20 & Full LLM reranking & 87.00 & 93.00 & 94.00 & 92.00 & 92.72 & 92.40 \\
\bottomrule
\end{tabular}%
}
\end{table*}

The persistent improvement in B5+ shows that accumulated retrieval experience remains useful when direct scores for the current query are sparse.
Thus, \method maintains a stable advantage over semantic retrieval even after the per-query reranking ratio has fallen to roughly one sixth of the candidate pool.

\subsection{Directly Scored versus Completed Memories}

Table~\ref{tab:component-ablation} separates the two sources of gain in \method.
Observed-only disables matrix completion and retains only memories with direct LLM relevance scores.
It improves over semantic retrieval by 1.36--3.83\%, demonstrating that the sampled direct scores provide a strong relevance signal.
Re-enabling completed scores adds a further 0.78--2.79\% in every configuration.
The largest completion contribution occurs at $r=8$, Top-10, where completed-score-selected memories add 2.79\% on top of the 3.83\% gain provided by directly scored memories.

\begin{table*}[t]
\centering
\caption{Component ablation separating directly scored and completed-score-selected memories. ``Observed in context'' is the mean number of selected memories with direct LLM relevance scores. The Observed-only variant disables completion; Mixed re-enables memories selected by completed scores.}
\label{tab:component-ablation}
\resizebox{\textwidth}{!}{%
\begin{tabular}{cccccccc}
\toprule
Rank $r$ & Top-$K$ & Observed in context & Semantic & Observed-only & Mixed (\method) & Observed $-$ Semantic & Mixed $-$ Observed \\
\midrule
2 & 10 & 4.59 & 82.21\% & 85.84\% & 87.27\% & +3.64\% & +1.43\% \\
8 & 10 & 5.04 & 82.21\% & 86.04\% & \textbf{88.83\%} & +3.83\% & \textbf{+2.79\%} \\
2 & 20 & 7.08 & 87.08\% & 88.51\% & \textbf{90.06\%} & +1.43\% & +1.56\% \\
8 & 20 & 7.49 & 87.08\% & 88.44\% & 89.22\% & +1.36\% & +0.78\% \\
\bottomrule
\end{tabular}%
}
\end{table*}

Together, the three variants provide a direct decomposition of the end-to-end improvement.
Semantic and Mixed use the same context budget, establishing the overall benefit of completion-based reranking at fixed $K$.
Observed-only and Mixed use the same directly acquired LLM scores, but only Mixed admits memories selected through imputed scores.
The consistently positive 0.78--2.79\% Mixed--Observed difference therefore shows that imputation itself improves the selected context, rather than the overall gain arising only from the small set of true LLM scores.

\section{Discussion}
\label{sec:discussion}

\subsection{From Content Memory to Retrieval Memory}

Agent-memory research usually asks what an agent should store and how stored content should be represented.
Our formulation adds a second persistent object: \emph{retrieval memory}, the accumulated evidence about which memories have been useful for which information needs.
Content memory answers ``what has the agent experienced?''
Retrieval memory answers ``what has the agent learned about accessing that experience?''
This distinction changes reranking scores from disposable inference traces into state that can improve future retrieval.

\subsection{Why the Method Is More Than a Cache}

A cache reuses the result of an identical or near-duplicate query~\cite{bang-2023-gptcache}.
\method instead seeks to transfer across distinct queries through shared memory biases, query biases, and latent interactions.
It can therefore benefit a new query even when the exact query has never appeared, provided that its sampled anchors connect it to structure learned from previous columns.
Candidate overlap helps, but exact repetition is not required.

\subsection{From Candidate Reranking to Search Expansion}

The retrieval-experience matrix also makes semantic retrieval a soft rather than irreversible boundary.
If a persistent memory has acquired row parameters from earlier queries but is absent from the current candidate set, its white cell in Figure~\ref{fig:earm-pipeline} can still be completed for the new query.
Ranking these estimates together with the semantic candidates expands the search domain and allows historically supported memories to enter the answer context even when cosine similarity misses a temporal, causal, or complementary relation.
The fixed-pool experiments isolate the reranking effect by setting $D_t=C_t$; the expanded domain $D_t=C_t\cup E_t$ follows from the same completion model without changing the relevance-scoring budget.

\subsection{What Must Be True for Amortization to Work}

The main hidden assumption is that query--memory relevance contains persistent cross-query structure.
This structure can arise from recurring user interests, memories with stable importance, repeated entities and events, or common types of information need.
If every query concerns an unrelated memory subset, if memory identities are unstable, or if the meaning of stored memories changes rapidly, historical scores offer little leverage.
The stage-wise results provide evidence that useful structure persists into the lowest-observation stage: with only 17.5\% of candidates directly scored, both completion ranks remain more accurate than semantic retrieval.

\section{Limitations}
\label{sec:limitations}

The current framework has three limitations.
First, its budget schedule is predetermined by query index and does not react to model uncertainty, distribution shift, or cold-start memories.
Second, matrix completion assumes reusable low-dimensional structure; highly heterogeneous query streams can violate this assumption.
Third, pointwise LLM relevance scores can be noisy, miscalibrated, and prompt-sensitive, so completion can propagate reranker errors rather than only reduce cost.

\section{Conclusion}
\label{sec:conclusion}

Long-lived agents should not repeatedly forget what their own retrieval decisions have revealed.
We presented \method, a framework that treats LLM relevance scores as retrieval experience, accumulates them in a causal query--memory matrix, and uses a latent accessibility map to rank candidates that were not directly scored.
Experiments on LoCoMo show that mixed observed-and-completed rankings improve end-to-end answer accuracy over semantic retrieval, retain their advantage at a 17.5\% direct-scoring ratio, and recover additional gains from completed-score-selected memories beyond directly scored anchors alone.
More broadly, a useful long-term memory system should not only remember more; it should become better at retrieving what it remembers.

\ifdefined\ARXIVVERSION
\bibliographystyle{unsrtnat}
\else
\bibliographystyle{IEEEtran}
\fi
\bibliography{references}

\end{document}